\documentclass[final]{cvpr}

\usepackage{times}
\usepackage{epsfig}
\usepackage{graphicx}
\usepackage{amsmath}
\usepackage{amssymb}
\usepackage{multirow}
\usepackage{booktabs}
\usepackage{color}

\usepackage[pagebackref=true,breaklinks=true,colorlinks,bookmarks=false]{hyperref}

\def\cvprPaperID{6754} % *** Enter the CVPR Paper ID here
\def\confYear{CVPR 2021}

\begin{document}

%%%%%%%%% TITLE
\title{Goal-Oriented Gaze Estimation for Zero-Shot Learning}

\author{Yang Liu$^{*\,1}$, Lei Zhou$^{*\,1}$, Xiao Bai$^{1}$, Yifei Huang$^{2}$, Lin Gu$^{3,2}$, Jun Zhou$^{4}$, Tatsuya Harada$^{2,3}$\\
	$^1$Beihang University, $^2$The University of Tokyo, $^3$RIKEN AIP, Tokyo, Japan, $^4$Griffith University\\
	{\tt\small \{wickerboy, leizhou, baixiao\}@buaa.edu.cn}, {\tt\small hyf@iis.u-tokyo.ac.jp}, {\tt\small lin.gu@riken.jp}\\ {\tt\small jun.zhou@griffith.edu.au}, {\tt\small harada@mi.t.u-tokyo.ac.jp}}

\maketitle

\footnotetext[1]{Equal contribution. }

%%%%%%%%% ABSTRACT
\begin{abstract}
   Zero-shot learning (ZSL) aims to recognize novel classes by transferring semantic knowledge from seen classes to unseen classes. Since semantic knowledge is built on attributes shared between different classes, which are highly local, strong prior for localization of object attribute is beneficial for visual-semantic embedding. Interestingly, when recognizing unseen images, human would also automatically gaze at regions with certain semantic clue. Therefore, we introduce a novel goal-oriented gaze estimation module (GEM) to improve the discriminative attribute localization based on the class-level attributes for ZSL. We aim to predict the actual human gaze location to get the visual attention regions for recognizing a novel object guided by attribute description. Specifically, the task-dependent attention is learned with the goal-oriented GEM, and the global image features are simultaneously optimized with the regression of local attribute features. Experiments on three ZSL benchmarks, i.e., CUB, SUN and AWA2, show the superiority or competitiveness of our proposed method against the state-of-the-art ZSL methods. The ablation analysis on real gaze data CUB-VWSW also validates the benefits and accuracy of our gaze estimation module. This work implies the promising benefits of collecting human gaze dataset and automatic gaze estimation algorithms on high-level computer vision tasks. The code is available at \url{https://github.com/osierboy/GEM-ZSL}.
\end{abstract}

%%%%%%%%% BODY TEXT
\section{Introduction}

With prior knowledge on seen classes, humans have a remarkable ability to recognize novel classes using shared and distinct attributes of both seen and unseen classes. Inspired by this cognitive competence, zero-shot learning (ZSL) was proposed as a challenging image classification setting to mimic the human cognitive process~\cite{lampert2009learning}. Given the semantic descriptions of both seen and unseen classes but only the training images of seen classes, ZSL aims to classify test images of unseen classes. 

Based on the classes that a model sees in the test phase, ZSL can be categorized into conventional or generalized setting. In conventional ZSL, the test images belong only to unseen classes. For the more practical and challenging generalized ZSL (GZSL) setting, the test images may belong to both seen and unseen classes. The semantic descriptions (attributes) are shared information between seen and unseen classes, which ensure the knowledge transferring. Early works~\cite{akata2013label,akata2015evaluation,frome2013devise,morgado2017semantically,romera2015embarrassingly} on ZSL build embedding between seen classes and their attributes. Then unseen classes are classified by the nearest neighbor search in the embedding space. These embedding based methods usually have a large bias towards seen classes under the GZSL setting, since the embedding is learned only by seen classes samples. To solve this problem, by leveraging the generative models~\cite{goodfellow2014generative,kingma2013auto,kingma2018glow}, many feature generation approaches~\cite{kumar2018generalized,xian2018feature,yu2019zero,schonfeld2019generalized,shen2020invertible,liu2020information} have been proposed to generate unseen classes, than convert ZSL into a conventional classification problem.

\begin{figure*}[t]
	\centering
	\includegraphics[width=1\linewidth]{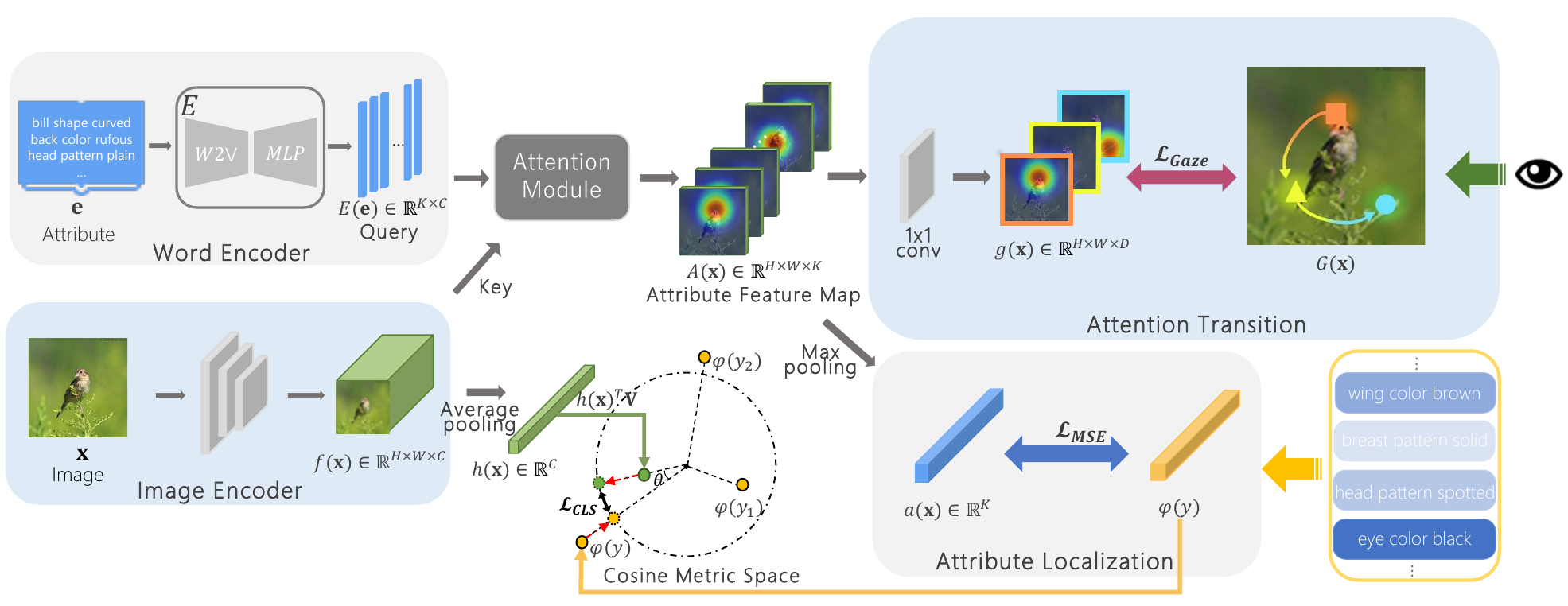}
	\caption{Illustration of the proposed method. Our GEM-ZSL is an end-to-end trainable model with two main parts, a gaze estimation module and a plain ZSL backbone. For the gaze estimation module, the projected word vector of each attribute by \textbf{Word Encoder} is first utilized as query to guide the localized attribute learning. The ground truth attributes $\mathbf{\varphi}(y)$ of specific seen class are simultaneously used as supervision in the \textbf{Attribute Localization} module. Then, an \textbf{Attention Transition} module is utilized to calibrate the attention regions by real gaze data with a designed gaze loss. Finally, the plain ZSL backbone learns the global image feature which is jointly optimized with the localized attribute, and the backbone accomplishes zero-shot recognition in a \textbf{Cosine Metric Space}.}
	\label{illustration}
\end{figure*}

Most of the existing embedding or feature generation based methods extract global features from pre-trained or end-to-end trainable models. However, only the global image features cannot effectively represent the fine-grained information between seen and unseen images, which is important for ZSL. More recently, attention based end-to-end models~\cite{xie2019attentive,zhu2019semantic,zhu2019generalized,xie2020region} have tried to exploit the semantic vector as guidance to learn more discriminative part features. However, they simply learn regions embedding of different attribute features but neglect the importance of discriminative attribute localization~\cite{sylvain2020locality,xu2020attribute}.

Very interestingly, when facing an unseen object with the guidance of an attribute description, humans are capable of paying attention to parts of the object with discriminative attributes, which is the gaze behavior. Karessli~\textit{et al.}~\cite{karessli2017gaze} proved that gaze data is useful in ZSL since they provide an effective prior that can naturally capture the localized discriminative attribute. Inspired by the human gaze mechanism, we propose a novel goal-oriented Gaze Estimation Method for Zero-Shot Learning (GEM-ZSL). As shown in Figure~\ref{illustration}, we first design an attribute description-oriented gaze estimation module (GEM) to learn different attribute regions. The GEM consists of three sub-modules, attention module (AM), attention transition (AT) module and attribute localization (AL) module. The AM is based on bilinear pooling which is widely used in the visual question answering task~\cite{fukui2016multimodal,yu2017multi}. We use the projected word vector of attributes as query to guide the learning process for the localization of the discriminative attribute. The ground truth attributes of specific seen class are simultaneously used as supervision by the mean squared error (MSE) of the AL module. Then an AT module is utilized to calibrate the attention regions by real gaze data (if available) with a designed gaze loss. In this way, we can learn the part feature with localized discriminative attribute that humans subconsciously pay attention to. Finally, the joint global features learned by an image encoder (IE), the local features and the class semantic embedding are used to learn a cosine metric space, which helps to reduce the intra-class variance and improve the recognition of unseen classes when compared to dot product similarity.

The contribution of this paper can be summarized as follows: 1. We propose a novel goal-oriented gaze estimation method to mimic the human cognitive process for recognizing unseen classes. With the guidance of attribute description, the proposed method can predict the human gaze that can be transformed to attribute attention for zero-shot recognition. 2. We demonstrate the effectiveness of our method for improving the localization of discriminative attributes, which further enhances the discrimination of the global features for ZSL task. 3. Comprehensive experiments over three ZSL benchmarks, {\it i.e.}, CUB, AWA2 and SUN, show that our method can achieve superior or competitive performance compared with the state-of-the-art ZSL methods. In addition, the quantitative and qualitative results on gaze estimation experiment also validate the effectiveness of our GEM.

%\begin{enumerate}
%	\item We propose a novel goal-oriented gaze estimation method to mimic the human cognitive process for recognizing unseen classes. With the guidance of attribute description, the proposed method can predict the human gaze that can be transformed to attribute attention for zero-shot recognition.
%	\item We demonstrate the effectiveness of our method for improving the localization of discriminative attributes, which further enhances the discrimination of the global features for ZSL task.
%	\item Comprehensive experiments over three ZSL benchmarks, {\it i.e.}, CUB, AWA2 and SUN, show that our method can achieve superior or competitive performance compared with the state-of-the-art ZSL methods. In addition, the quantitative and qualitative results on gaze estimation experiment also validate the effectiveness of our GEM.
%\end{enumerate}

%-------------------------------------------------------------------------
\section{Related Works}

\subsection{Early ZSL}

Early ZSL methods~\cite{akata2013label,frome2013devise,romera2015embarrassingly,akata2015evaluation,xian2016latent,xian2017zero,zhang2017learning} focus on learning a mapping between visual and semantic spaces to transfer semantic knowledge from seen classes to unseen classes. However, these methods usually achieve relatively unsatisfied results, since they adopt global features or exploit shallow models. In more recent, end-to-end deep models~\cite{morgado2017semantically,liu2018generalized,song2018transductive,liu2019attribute} achieve better performance. These methods constrain loss on the attributes of seen classes to allow learning of more discriminative global features. However, they neglect to focus on the parts of features which are intrinsically discriminative for ZSL.

\subsection{Part-based ZSL}
More relevant to this work is the recently part-based ZSL methods~\cite{ji2018stacked,xie2019attentive,zhu2019semantic,xie2020region,xu2020attribute} that utilize attribute descriptions as guidance to learn discriminative part features. These methods have achieved remarkable improvements on ZSL. In order to improve the GZSL, APN~\cite{xu2020attribute} applies calibrated stacking (CS)~\cite{chao2016empirical} in test phase to reduce the seen class scores by a constant factor, which can significantly improve the GZSL performance. RGEN~~\cite{xie2020region} designs a balance loss by pursuing the maximum response consistency among seen and unseen outputs in the training phase, which requires the attributes of unseen classes during training. Compared with these methods, our GEM-ZSL improves the localization of discriminative attributes by goal-oriented gaze estimation for the first time. In addition, only the attributes of seen classes are used in the model training.

\subsection{Gaze Estimation}
Since human gaze directly represents attention and thus the important spatial regions, many works try to estimate where humans will look given an image. Earlier works~\cite{harel2007graph,achanta2009frequency,cheng2014global} try to model saliency in the images that are likely to attract human attention. These methods typically use feature integration theory~\cite{treisman1980feature} to fuse various bottom-up cues. More recently researchers begin to use deep learning models for the estimation of human gaze~\cite{jiang2015salicon,pan2016shallow,wang2019revisiting}. However, these models only consider the bottom-up information which is not suitable for ZSL since they cannot focus on the fine-grained discriminative regions. Some newly proposed methods~\cite{huang2018predicting,huang2020mutual} leverage top-down task-specific cues for estimating human gaze. However these methods are typically designed for some specific tasks and need large amount of training data. In this work, to solve the ZSL task, we exploit the attribute description as guidance to design a novel goal-oriented gaze estimation module.

%-------------------------------------------------------------------------
\section{The Proposed Method}

In this section, we first define the problem setting, notations and then present the details of each module of our method. 

\subsection{Problem Setting and Notations}

ZSL aims to recognize novel classes by transferring semantic knowledge from seen classes ($\mathcal{Y}^S$) to unseen classes ($\mathcal{Y}^U$). The image spaces of seen and unseen classes can be defined as $\mathcal{X}= \mathcal{X}^S \cup \mathcal{X}^U$.  $\mathcal{S}=\{(\mathbf{x},y,\mathbf{\varphi}(y))|\mathbf{x}\in \mathcal{X}^S,y\in \mathcal{Y}^S,\mathbf{\varphi}(y)\in \mathcal{\phi}^S\}$ denotes the training set,  where $\mathbf{x}$ is an image in $\mathcal{X}^S$, $y$ is its class label which is available during training, and $\mathbf{\varphi}(y)\in \mathbb{R}^K$ is the class semantic embedding, {\it i.e.}, a class-level attribute vector annotated with $K$ different visual attributes. The unseen testing set is $\mathcal{U}=\{(\mathbf{x}^u, u,\mathbf{\varphi}(u))|\mathbf{x}^u\in \mathcal{X}^U, u \in \mathcal{Y}^U,\mathbf{\varphi}(u)\in \mathcal{\phi}^U\}$, where $u$ denote unseen class labels. The seen classes and unseen classes are disjoint, {\it i.e.}, $\mathcal{Y}^S\cap \mathcal{Y}^U=\emptyset$. Here, $\mathcal{\phi}=\mathcal{\phi}^S\cup \mathcal{\phi}^U$ is used to transfer information between seen and unseen classes. In the conventional ZSL, the task is to predict the label of images from unseen classes , {\it i.e.},  $\mathcal{X}^U \rightarrow \mathcal{Y}^U$. However, in more realistic and challenging setup of GZSL, the aim is to predict images from both seen and unseen classes, {\it i.e.},  $\mathcal{X} \rightarrow \mathcal{Y}^U\cup \mathcal{Y}^S$.

As shown in Figure~\ref{illustration}, our end-to-end trained model consists of two parts, Gaze Estimation Module (GEM) and a plain ZSL backbone. Specifically, GEM consists of Word Encoder (WE) module, Attention Module (AM), Attribute Localization (AL) module and Attention Transition (AT) module. The plain ZSL backbone is an Image Encoder (IE) with a Cosine Metric Space for the nearest neighbor search. Firstly , IE is used to extract the global feature of the image. Then, GEM learns the part feature with localized discriminative attribute that humans subconsciously pay attention to. Finally, the joint global and local feature and the class semantic embedding are used to learn a cosine metric space.

\subsection{Cosine Metric Learning}

\textbf{Visual feature extraction.} We leverage the Image Encoder (IE) implemented by a convolutional neural network to map the seen class image $\mathbf{x}$ into a feature representation $\textit{f}(\mathbf{x}) \in \mathbb{R}^{\textit{H} \times \textit{W} \times \textit{C}}$, where $\textit{H}$, $\textit{W}$ and $\textit{C}$ are the height, width and channel of the feature, respectively. Then, global average pooling is applied over the $\textit{H}$ and $\textit{W}$ to learn a global discriminative feature $h(\mathbf{x}) \in \mathbb{R}^\textit{C}$.

\textbf{Cosine similarity for classification.} 	We use a linear layer $ \mathbf{V} \in \mathbb{R}^{\textit{C} \times \textit{K}}$ to map the visual feature $h(\mathbf{x})$ into the semantic space. Different from previous work that~\cite{xu2020attribute} uses dot product to compute class logits of the projected visual feature and every class embedding, we consider the cosine similarity~\cite{gidaris2018dynamic,luo2018cosine} that can bound and reduce the variance of the neurons and thus result in models with better generalization capability~\cite{li2019rethinking}. The output of the cosine metric is scaled similarity score of the projected visual feature $h(\mathbf{x})$ and $y$-th class semantic embedding $\mathbf{\varphi}(y)$. Then we define our classification score function as
\begin{equation}
p(y|\mathbf{x})=\dfrac{\exp(\sigma \cos(h(\mathbf{x})^T\mathbf{V},\mathbf{\varphi}(y)))}{\sum\nolimits_{\hat{y} \in\mathcal{Y}^S }\exp(\sigma \cos(h(\mathbf{x})^T\mathbf{V},\mathbf{\varphi}(\hat{y})))}
\end{equation}
where $\sigma$ is the scaling factor. The classification loss $L_{CLS}$ is defined as
\begin{equation}
\mathcal{L}_{CLS}=-\log\dfrac{\exp(\sigma \cos(h(\mathbf{x})^T\mathbf{V},\mathbf{\varphi}(y)))}{\sum\nolimits_{\hat{y} \in\mathcal{Y}^S }\exp(\sigma \cos(h(\mathbf{x})^T\mathbf{V},\mathbf{\varphi}(\hat{y})))}
\end{equation}

We find empirically that using the image feature and class semantic embedding normalization in the cosine similarity helps to reduce the intra-class variance and improve the accuracy of unseen classes when compared to dot product similarity.

\subsection{Gaze Estimation Module}

\begin{figure}[t]
	\centering
	\includegraphics[width=1\linewidth]{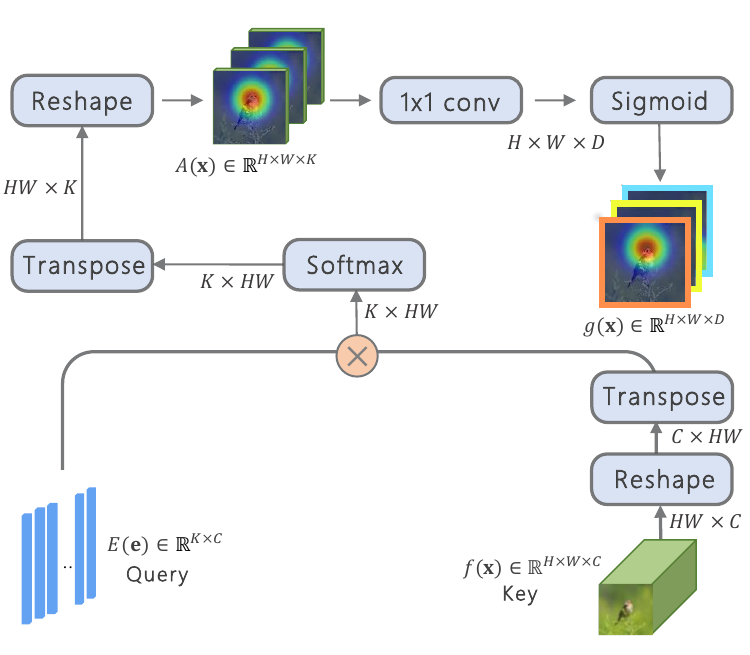}
	\caption{The structure of attention module and attention transition module. It is based on the bilinear pooling mechanism. The feature map (key) and word vector (query) are shown as the shape of their tensor, {\it i.e.} $\textit{H} \times \textit{W} \times \textit{C}$ for key and $\textit{K} \times \textit{C}$ for query. $\otimes$  denotes matrix multiplication. The softmax operation is performed on each row.}
	\label{attention}
\end{figure}

Although cosine similarity shows better performance than the dot product similarity, the global features learned from cosine-based image encoder may still be biased to seen classes and weaken the discriminative attributes information. We propose a gaze estimation module (GEM) to learn the part feature with localized discriminative attribute that humans subconsciously pay attention to.

\textbf{Word encoder.} To learn local features, the attribute semantic vectors $\mathbf{e}=\{\mathbf{e}_k\}^K_{k=1}$ are utilized to guide the learning of the localized discriminative attribute, where $\mathbf{e}_k$ denotes the average GloVe~\cite{pennington2014glove} representations of words in the $k$-th attribute, e.g., ``brown eye" and ``plain head". Then, a single hidden layer MLP is used to convert the attribute word vector $\mathbf{e}$ into visual attribute feature $E(\mathbf{e}) \in \mathbb{R}^{\textit{K} \times \textit{C}}$.

\textbf{Attention module.} To acquire part features with localized discriminative attribute, we design an attribute descriptions-oriented attention module to learn the different attribute regions. As shown in Figure~\ref{attention}, AM is based on bilinear pooling mechanism which is widely used in the visual question answering task~\cite{fukui2016multimodal,yu2017multi}. We use the projected word vector of attributes $E(\mathbf{e})$ as a query to guide learning localized discriminative attribute. The inputs of AM are query $E(\mathbf{e}) \in \mathbb{R}^{\textit{K} \times \textit{C}}$ and key $\textit{f}(\mathbf{x}) \in \mathbb{R}^{\textit{H} \times \textit{W} \times \textit{C}}$. Firstly, AM reshapes the dimension of key into $\textit{HW} \times \textit{C}$ and transposes it into $\textit{C} \times \textit{HW}$. Then, AM performs matrix multiplication and softmax operation on query and key. The $k$-th attribute query captures related attribute part of image and produces $A(\mathbf{x})^k$ that contains the $k$-th localized attribute information. After transposing and reshaping the feature, the attribute feature map $A(\mathbf{x}) \in \mathbb{R}^{\textit{H} \times \textit{W} \times \textit{K}}$ is acquired. The localized discriminative attribute is expected to concentrate to a peak region rather than disperses on other locations. We use Distance loss~\cite{zheng2017learning} on each attention map $A(\mathbf{x})^k$ to constrain the discriminative attribute localization, such that
\begin{equation}
\mathcal{L}_{Dis}=\sum_{k=1}^K\sum_{i=1}^H\sum_{j=1}^W A(\mathbf{x})^k_{i,j}(\left\|i-\tilde{i}\right\|^2+\left\| j-\tilde{j} \right\|^2)
\end{equation}
where $(\tilde{i},\tilde{j})=\arg\max_{i,j}A(\mathbf{x})^k$ denotes the coordinate for the maximum value in $A(\mathbf{x})^k$. We can learn a more concentrated attention map by this objective function.

\textbf{Attribute localization.} We predict the location of the attributes response value $a(\mathbf{x})$ by global max pooling over the $\textit{H}$ and $\textit{W}$ on $A(\mathbf{x})$. Then $a(\mathbf{x})$ is optimized by the Mean Square Error (MSE) with the supervision of the ground truth attributes $\mathbf{\varphi}(y)$
\begin{equation}
\mathcal{L}_{MSE}=\left\|a(\mathbf{x})-\mathbf{\varphi}(y)\right\|^2_2
\end{equation}
where $y$ is the ground truth class. We make the local features correspond to discriminative attributes by minimizing the MSE loss, which improves the discriminative attribute localization.

\begin{figure}[t]
	\centering
	\includegraphics[width=1\linewidth]{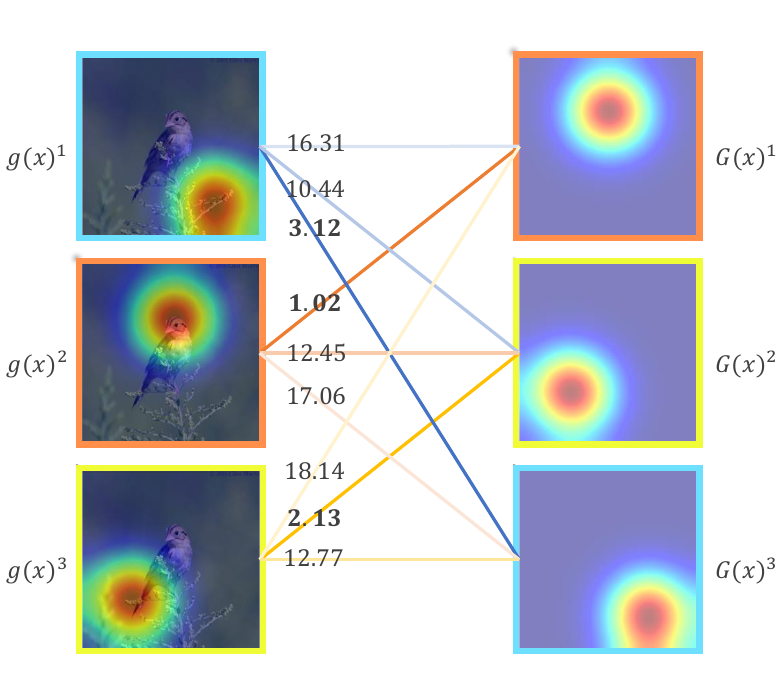}
	\caption{The predicted gaze map $g(\mathbf{x})$ and human gaze $G(\mathbf{x})$ matching method. We utilize the Hungarian algorithm which calculates the $l1$ distance between $g(\mathbf{x})$ and $G(\mathbf{x})$, and matches them one-to-one.}
	\label{hungarian}
\end{figure}

\textbf{Attention transition.} To acquire the part feature with discriminative attribute of human gaze, we propose an AT module to convert attribute feature map $A(\mathbf{x}) \in \mathbb{R}^{\textit{H} \times \textit{W} \times \textit{K}}$ to gaze map $g(\mathbf{x}) \in \mathbb{R}^{\textit{H} \times \textit{W} \times \textit{D}}$, where $\textit{D}$ represents the number of gaze heatmaps. We have true human gaze $G(\mathbf{x}) \in \mathbb{R}^{\textit{H} \times \textit{W} \times \textit{D}}$ of image $\mathbf{x}$ to enable supervised learning. The AT is implemented by a learnable $1 \times 1$ convolution operation and a sigmoid function, which fuses different attributes information and converts them into a region of human attention. To optimize the gaze map that incorporates the discriminative attribute regions, we propose a new gaze loss. Firstly, we use the Hungarian algorithm (Figure~\ref{hungarian}) to match $G(\mathbf{x})$ and $g(\mathbf{x})$, where the matched gaze map and its ground truth are termed $\tilde{g}(\mathbf{x})$ and $\tilde{G}(\mathbf{x})$, respectively. Then we use a binary cross-entropy loss across all the pixels as
\begin{equation}
\begin{split}
\mathcal{L}_{Gaze}=-\frac{1}{\textit{D} \times \textit{H} \times \textit{W}  }\sum_{d=1}^D\sum_{i=1}^H\sum_{j=1}^W\{\tilde{G}(\mathbf{x})^d_{i,j}\log(\tilde{g}(\mathbf{x})^d_{i,j})\\
+(1-\tilde{G}(\mathbf{x})^d_{i,j})\log(1-\tilde{g}(\mathbf{x})^d_{i,j}) \}
\end{split}
\end{equation}

Our full model optimizes the CNN backbone and the overall loss of the proposed model is defined as
\begin{equation}
\mathcal{L}=\mathcal{L}_{CLS}+\lambda_1 \mathcal{L}_{Dis}+\lambda_2 \mathcal{L}_{MSE}+\lambda_3 \mathcal{L}_{Gaze}
\end{equation}
where $\lambda_1$, $\lambda2$ and $\lambda_3$ are hyper-parameters for Distance loss, Mean Square Error loss and Gaze loss, respectively. Particularly, $\lambda_3=0$ when gaze ground truth is not available. The joint training improves the representation of the discriminative attribute that is critical for zero-shot generalization.

\subsection{Zero-Shot Recognition}

After training of the full model, we use the learned cosine metric space for zero-shot recognition. The test image $\mathbf{x}$ is embedded to the cosine metric space by the visual-semantic embedding layer, and then the classifier searches for the class embedding $\mathbf{\varphi}(\hat{u})$ with the highest compatibility via
\begin{equation}
\hat{u}= \mathop{\arg\max}_{u \in \mathcal{Y}^U}  \cos(h(\mathbf{x})^T\mathbf{V},\mathbf{\varphi}(u))
\end{equation}

For the GZSL setting, the test images may belong to both seen and unseen classes. Since there are only seen classes during training phase, the predicted results of GZSL will have a large bias towards seen classes~\cite{chao2016empirical}. To mitigate this problem, we apply calibrated stacking (CS)~\cite{chao2016empirical} to reduce the seen class scores by a calibration factor $\gamma$. Specifically, the GZSL classifier is defined as
\begin{equation}
\hat{y}= \mathop{\arg\max}_{\tilde{y} \in \mathcal{Y}^U \cup \mathcal{Y}^S} (\sigma \cos(h(\mathbf{x})^T\mathbf{V},\mathbf{\varphi}(\tilde{y}))-\gamma \mathbb{I}[\tilde{y} \in \mathcal{Y}^S])
\end{equation}
where $\mathbb{I}=1$ if $\tilde{y}$ is a seen class and 0 otherwise.

\subsection{Implementation Details}
The proposed GEM-ZSL is an end-to-end trainable model. The Image Encoder is ResNet101~\cite{he2016deep} pretrained on ImageNet~\cite{deng2009imagenet}. The SGD~\cite{bottou2010large} optimizer is adopted in the model training. The momentum is set to $0.9$, and the weight decay is $10^{-5}$. The learning rate is $10^{-3}$. We set $\lambda_1$ to $0.2$ and $\lambda_2$ to $1.0$. When we have gaze attention, $\lambda_3$ is set to $0.1$. The factor $\gamma$ is set to $3.5$ for AWA2 and $0.7$ for CUB and SUN. We use an episode-based training method to sample $M$ categories and $N$ images for each category in a mini-batch, we iterate $300$ batches for each epoch, and train the model $20$ epochs. We set $M=16$ and $N=2$ for all three datasets.

%------------------------------------------------------------------------
\begin{table*}[t]
	\centering
	\caption{Results ($\%$) of the state-of-the-art ZSL and GZSL. The first part is non end-to-end methods, the second part is  feature generation methods and the third part is end-to-end methods. The best and the second best results are marked in red and blue, respectively.}
	\begin{center}
		\scalebox{0.9}{
		\begin{tabular}{c|c|c|c|c|c|c|c|c|c|c|c|c}
			\toprule[1.5pt]
			\multirow{3}{*}{Methods}
			&\multicolumn{4}{c|}{CUB}&\multicolumn{4}{c|}{SUN}&\multicolumn{4}{c}{AWA2}\cr\cline{2-13}
			&\multicolumn{1}{c|}{ZSL}&\multicolumn{3}{c|}{GZSL}&\multicolumn{1}{c|}{ZSL}&\multicolumn{3}{c|}{GZSL}&\multicolumn{1}{c|}{ZSL}&\multicolumn{3}{c}{GZSL}\cr\cline{2-13}
			&\textbf{T1}&\textbf{U}&\textbf{S}&\textbf{H}&\textbf{T1}&\textbf{U}&\textbf{S}&\textbf{H}&\textbf{T1}&\textbf{U}&\textbf{S}&\textbf{H}\cr
			\hline
			\hline
			PSR(CVPR'18)~\cite{annadani2018preserving} & 56.0 &	24.6 & 54.3 & 33.9 & 61.4 &	20.8 &	37.2 &	26.7 &	63.8 & 20.7 & 73.8 & 32.3
			\cr
			RN(CVPR'18)~\cite{sung2018learning} & 55.6 &	38.1 &	61.1 &	47.0 &	- &	- &	- &	- &	64.2 &	30.0 &	\textcolor[rgb]{0,0,1}{93.4} &	45.3
			\cr
			SP-AEN(CVPR'18)~\cite{chen2018zero} & 55.4 & 	34.7 &	70.6 &	46.6 &	59.2 &	24.9 &	38.6 &	30.3 &	- &	- &	- &	- 
			\cr
			IIR(ICCV'19)~\cite{cacheux2019modeling} & 63.8 &	55.8 &	52.3 &	53.0 &	63.5 &	47.9 &	30.4 &	36.8 &	67.9 &	48.5 &	83.2 &	61.3
			\cr
			TCN(ICCV'19)~\cite{jiang2019transferable} & 59.5 &	52.6 &	52.0 &	52.3 &	61.5 &	31.2 &	37.3 &	34.0 &	71.2 &	61.2 &	65.8 &	63.4
			\cr
			E-PGN(CVPR'20)~\cite{yu2020episode} & 72.4 & 52.0 &	61.1 &	56.2 &	- &	- &	- &	- &	73.4 &	52.6 &	83.5 &	64.6
			\cr
			DAZLE(CVPR'20)~\cite{huynh2020fine} & 65.9 & 56.7 &	59.6 &	58.1 &	- &	52.3 &	24.3 &	33.2 &	- &	60.3 &	75.7 &	67.1
			\cr
			\hline
			\hline
			f-CLSWGAN(CVPR'18)~\cite{xian2018feature} & 57.3 &	43.7 &	57.7 &	49.7 &	60.8 &	42.6 &	36.6 &	39.4 &	- &	- &	- &	- 
			\cr
			cycle-CLSWGAN(ECCV'18)~\cite{felix2018multi} & 58.4 &	45.7 &	61.0 &	52.3 &	60.0 &	49.4 &	33.6 &	40.0 &	- &	- &	- &	- 
			\cr
			CADA-VAE(CVPR'19)~\cite{schonfeld2019generalized} & - &	51.6 &	53.5 &	52.4 &	-	& 47.2 &	35.7 &	40.6 &	- &	55.8 &	75.0 &	63.9
			\cr
			OCD-CVAE(CVPR'20)~\cite{keshari2020generalized} & 60.3 &	44.8 &	59.9 &	51.3 &	63.5 &	44.8 &	\textcolor[rgb]{0,0,1}{42.9} &	43.8 &	71.3 &	59.5 &	73.4 &	65.7 
			\cr
			RFF-GZSL(1-NN)(CVPR'20)~\cite{han2020learning} & - &	50.6 &	79.1 &	61.7 &	- &	\textcolor[rgb]{1,0,0}{56.6} &	42.8 &	\textcolor[rgb]{0,0,1}{48.7}&	- &	- &	- &	- 
			\cr
			IZF(ECCV'20)~\cite{shen2020invertible} & 67.1 &	52.7 &	68.0 &	59.4 &	\textcolor[rgb]{1,0,0}{68.4} &	\textcolor[rgb]{0,0,1}{52.7} &	\textcolor[rgb]{1,0,0}{57.0} &	\textcolor[rgb]{1,0,0}{54.8} &	\textcolor[rgb]{1,0,0}{74.5} &	60.6 &	77.5 &	68.0 
			\cr
			LsrGAN(ECCV'20)~\cite{vyas2020leveraging} & 60.3 &	48.1 &	59.1 &	53.0 &	62.5 &	44.8 &	37.7 &	40.9 &	- &	- &	- &	- 
			\cr
			\hline
			\hline
			QFSL(CVPR'18)~\cite{song2018transductive} & 58.8 &	33.3 &	48.1 &	39.4 &	56.2 &	30.9 &	18.5 &	23.1 &	63.5 &	52.1 &	72.8 &	60.7
			\cr
			LDF(CVPR'18)~\cite{li2018discriminative} & 67.5 &	26.4 &	\textcolor[rgb]{1,0,0}{81.6} &	39.9 &	- &	- &	- &	- &	- &	- &	- &	- 
			\cr
			SGMA(NeurIPS'19)~\cite{zhu2019semantic} & 71.0 &	36.7 &	71.3 &	48.5 &	- &	- &	- &	- &	- &	- &	- &	- 
			\cr
			AREN(CVPR'19)~\cite{xie2019attentive} & 71.8 &	63.2 &	69.0 &	66.0 &	60.6 &	40.3 &	32.3 &	35.9 &	67.9 &	54.7 &	79.1 &	64.7
			\cr
			LFGAA(ICCV'19)~\cite{liu2019attribute} & 67.6 &	36.2 &	\textcolor[rgb]{0,0,1}{80.9} &	50.0 &	61.5 &	18.5 &	40.0 &	25.3 &	68.1 &	27.0 &	\textcolor[rgb]{1,0,0}{93.4} &	41.9
			\cr
			DVBE(CVPR'20)~\cite{min2020domain} & - &	64.4 &	73.2 &	\textcolor[rgb]{0,0,1}{68.5} &	- &	44.1 &	41.6 &	42.8 &	- &	62.7 &	77.5 &	69.4
			\cr
			RGEN(ECCV'20)~\cite{xie2020region} & \textcolor[rgb]{0,0,1}{76.1} &	60.0 &	73.5 &	66.1 &	\textcolor[rgb]{0,0,1}{63.8} &	44.0 &	31.7 &	36.8 &	\textcolor[rgb]{0,0,1}{73.6} &	\textcolor[rgb]{1,0,0}{67.1} &	76.5 &	\textcolor[rgb]{1,0,0}{71.5} 
			\cr
			APN(NeurIPS'20)~\cite{xu2020attribute} & 72.0 &	\textcolor[rgb]{1,0,0}{65.3} &	69.3 &	67.2 &	61.6 &	41.9 &	34.0 &	37.6 &	68.4 &	56.5 &	78.0 &	65.5
			\cr
			GEM-ZSL(Ours) & \textcolor[rgb]{1,0,0}{77.8} & \textcolor[rgb]{0,0,1}{64.8} & 77.1 & \textcolor[rgb]{1,0,0}{70.4} & 62.8 & 38.1 & 35.7 & 36.9 & 67.3 & \textcolor[rgb]{0,0,1}{64.8} & 77.5 & \textcolor[rgb]{0,0,1}{70.6} \cr
			\bottomrule[1.5pt]
		\end{tabular}}
	\end{center}
	\label{gzsl}
\end{table*}

\begin{table}[t]
	\centering
	\caption{Results ($\%$) of ZSL and GZSL ablation study on CUB, SUN and AWA2. The baseline is the Image Encoder and dot product distance with cross-entropy loss. We analyzed the performance of each module of our model.}
	\begin{center}
		\begin{tabular}{c|c|c|c|c|c|c}
			\toprule[1.5pt]
			\multirow{2}{*}{Methods}&
			\multicolumn{2}{c|}{CUB}&\multicolumn{2}{c|}{SUN}&\multicolumn{2}{c}{AWA2}\cr\cline{2-7}
			&\textbf{T1}&\textbf{H}&\textbf{T1}&\textbf{H}&\textbf{T1}&\textbf{H}\cr
			\hline
			Baseline & 67.8 & 61.5 & 54.7 & 30.9 & 63.5 & 62.7 \cr
			$+\mathcal{L}_{MSE}$ & 69.9 & 63.1 & 56.1 & 31.5 & 64.3 & 64.9 \cr
			$+\mathcal{L}_{Dis}$ & 70.5 & 64.9 & 56.9 & 31.7 & 64.6 & 65.2 \cr
			$+cos$ & \textbf{77.8} & \textbf{70.4} & \textbf{62.8} & \textbf{36.9} & \textbf{67.3} & \textbf{70.6} \cr
			\bottomrule[1.5pt]
		\end{tabular}
	\end{center}
	\label{ablation}
\end{table}

\section{Experiments}
We evaluated our framework on three widely used zero-shot learning benchmark datasets, including CUB-200-2011 (CUB)~\cite{welinder2010caltech}, SUN attribute (SUN)~\cite{patterson2012sun} and Animals with Attributes 2 (AWA2)~\cite{xian2018zero}. The pre-defined attributes on each dataset were used as the semantic descriptors. Moreover, we adopted the Proposed Split (PS)~\cite{xian2018zero} to divide all classes into seen and unseen classes on each dataset.

The performance of ZSL is evaluated by average per-class Top-1 (\textbf{T1}) accuracy. In GZSL, since the test set is composed of seen and unseen images, the Top-1 accuracy evaluated respectively on seen classes, denoted as \textbf{S}, and unseen classes, denoted as \textbf{U}. Their harmonic mean, defined as $\textbf{H} = (2 \times \textbf{S} \times \textbf{U})/(\textbf{S} + \textbf{U})$~\cite{xian2018zero}, are used to evaluate the performance of GZSL.

\subsection{Comparison with the State-of-the-Art}
We selected recent state-of-the-art ZSL methods for comparison, which include methods without end-to-end training such as PSR~\cite{annadani2018preserving}, RN~\cite{sung2018learning}, SP-AEN~\cite{chen2018zero}, IIR~\cite{cacheux2019modeling}, TCN~\cite{jiang2019transferable}, E-PGN~\cite{yu2020episode}, DAZLE~\cite{huynh2020fine}, f-CLSWGAN~\cite{xian2018feature}, cycle-CLSWGAN~\cite{felix2018multi}, CADA-VAE~\cite{schonfeld2019generalized}, OCD-CVAE~\cite{keshari2020generalized}, RFF-GZSL~\cite{han2020learning}, IZF~\cite{shen2020invertible}, and LsrGAN~\cite{vyas2020leveraging}, where the last seven methods are feature generation based models, and end-to-end methods QFSL~\cite{song2018transductive}, LDF~\cite{li2018discriminative}, SGMA~\cite{zhu2019semantic}, AREN~\cite{xie2019attentive}, LFGAA~\cite{liu2019attribute}, DVBE~\cite{min2020domain}, RGEN~\cite{xie2020region}, and APN~\cite{xu2020attribute}.

Table~\ref{gzsl} shows the results of different methods on three datasets. Our GEM-ZSL achieves competitive performance compared with the state-of-the-art methods. On CUB dataset, GEM-ZSL outperforms all the compared methods with a large margin for both ZSL and GZSL. Since CUB is a more challenging fine-grained dataset which requires local discriminative attributes, the results prove the effectiveness of our localized attribute learning model. For AWA2 dataset, our GEM-ZSL can also achieve competitive result which is only slightly lower than RGEN. However, the balance loss of RGEN requires the attributes of unseen classes during training which is not used in our GEM-ZSL. Without the balance loss, the harmonic mean for RGEN on AWA2 will dramatically decrease to 14.7\%. On SUN dataset, the feature generation based model significantly outperforms the other methods. As SUN dataset contains more than 700 categories, the generative model can bring more features for generalization to unseen classes. Compared with the other non-generation based methods, the performance of our GEM-ZSL is competitive.

\subsection{Ablation Study}
\textbf{Component analysis.} We conducted ablation experiments to verify the effectiveness of the proposed modules. Table~\ref{ablation} shows the influence of each model component. Firstly, we train a baseline model that contains Image Encoder with cross-entropy loss. We use the dot product to compute class logits of the projected visual feature and every class embedding. Then, we add gaze estimation module, mean square error loss and distance loss functions gradually. Finally, the cosine similarity is added to the model. We can see that our proposed GEM improves the Top-1 accuracy (\textbf{T1}) of ZSL over the baseline consistently by $10.0\%$ (CUB), $8.1\%$ (SUN), $3.8\%$ (AWA2), and the harmonic mean accuracy (\textbf{H}) of GZSL over the baseline by $8.9\%$ (CUB), $6.0\%$ (SUN), $7.9\%$ (AWA2) respectively. Remarkably, when cosine distance is used to measure the similarity of the projected visual feature and class semantic embedding, the performance of our model can be greatly improved. Since the cosine metric can bound and reduce the variance of the neurons and thus result in models with better generalization capability~\cite{li2019rethinking}.

\begin{table}[t]
	\centering
	\caption{Results ($\%$) of ZSL ablation study for attention transition with gaze loss.}
	\begin{center}
		\begin{tabular}{c|c|c|c|c|c}
			\toprule[1.5pt]
			Methods&Set 1&Set 2&Set 3&Set 4&Avg.\cr
			\hline
			GEM w/o $\mathcal{L}_{Gaze}$  & 43.5& \textbf{45.5}& 42.1& 40.7 & 42.9 \cr
			GEM w $\mathcal{L}_{Gaze}$ & \textbf{44.9} & 45.2 & \textbf{43.6} & \textbf{41.2}& \textbf{43.7} \cr
			\bottomrule[1.5pt]
		\end{tabular}
	\end{center}
	\label{gaze}
\end{table}

\begin{figure*}[t]
	\centering
	\includegraphics[width=1\linewidth]{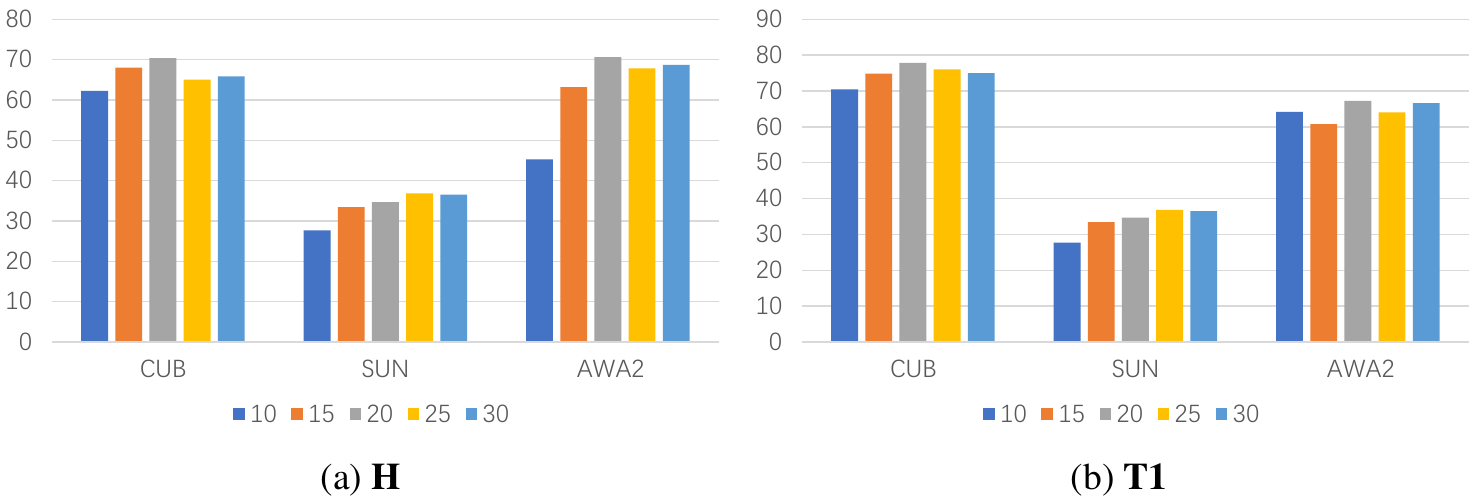}
	\caption{The influence of $\sigma$ with different values of ZSL (\textbf{T1}) and GZSL (\textbf{H}) results ($\%$) on CUB, SUN and AWA2.}
	\label{sigma}
\end{figure*}

To verify the influence of attention transition, we used CUB-VWSW gaze dataset with captured gaze points~\cite{karessli2017gaze}. We performed Gaussian blur on the gaze points to obtain the gaze heatmap as gaze ground truth. Then we randomly selected two categories for each bird family as unseen categories. The other categories were used as seen categories. All seen categories were used to train the model with gaze loss, and the unseen categories were used as a test set to verify model performance. We repeated the above operation for four times and took the average value of \textbf{T1} to ensure robustness. Table~\ref{gaze} shows the influence of AT with gaze loss,  and we can see that AT improves the ZSL accuracy.

\begin{figure}[t]
	\centering
	\includegraphics[width=1\linewidth]{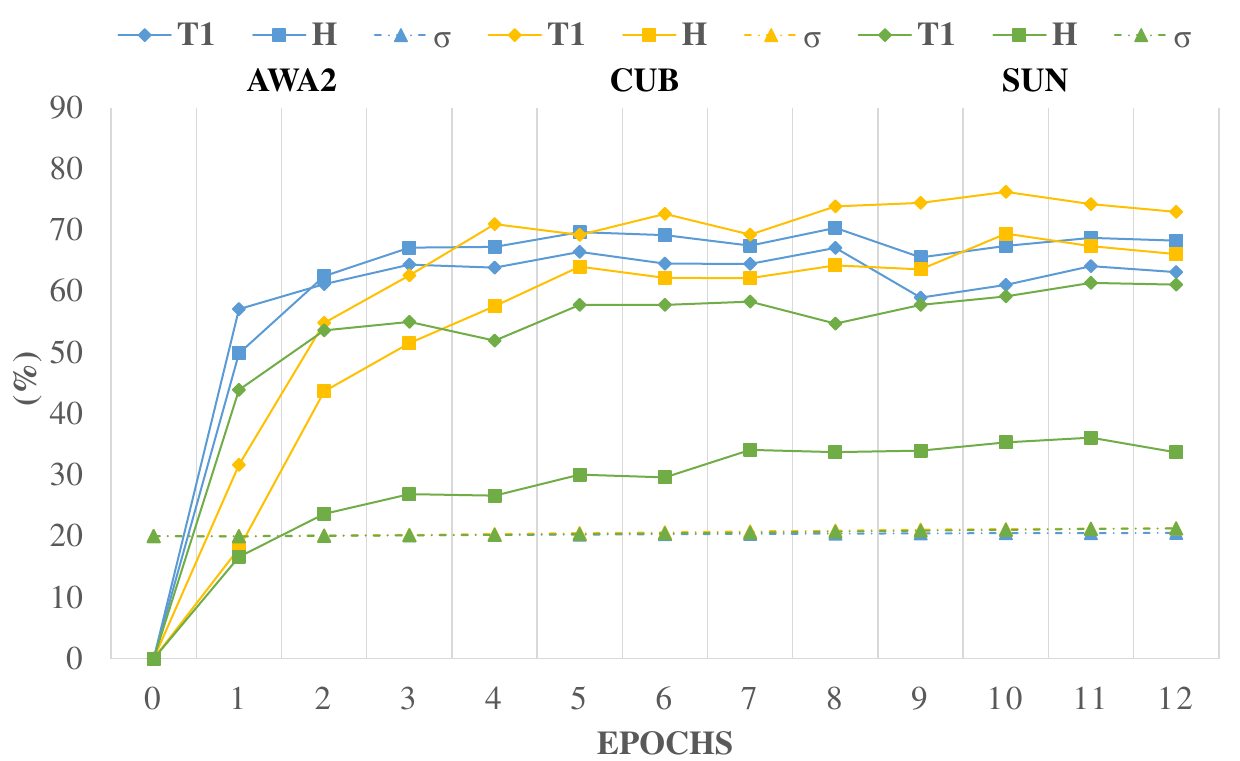}
	\caption{The influence of learnable $\sigma$ on ZSL (T1) and GZSL (H) results ($\%$). $\sigma$ is set to 20. During the model training, the performance increases with a slight fluctuation of $\sigma$.}
	\label{sigmatraining}
\end{figure}

\textbf{Effect of scaling factor $\sigma$.} Figure~\ref{sigma} shows the results of \textbf{H} and \textbf{T1}  when varying $\sigma$ over $\{10, 15, 20, 25, 30\}$ under GZSL/ZSL for our model. Our method achieves the best performance on CUB and AWA2 when $\sigma$ is 20. The best results are reached for SUN when $\sigma$ is 25. We further explore the performance of the model when $\sigma$ is a learnable parameter. Figure~\ref{sigmatraining} shows the relationship between model performance and $\sigma$ during the model training. We initialize $\sigma$ to 20, and it changes slightly with the iteration of training. Therefore, we fixed $\sigma$ to 20 for CUB and AWA2 and 25 for SUN in our experiments.
\begin{table}[t]
	\centering
	\caption{Influence of training method on GZSL results (\%). $\mathcal{R}$ represents random sampling training method with mini-batch of 64, $\mathcal{E}$ represents episode-based training method.}
	\begin{center}
		\scalebox{0.88}{
			\begin{tabular}{c | c c | c c c}
				\toprule[1.5pt]
				Training Method&$M$-way&$N$-shot&CUB&SUN&AWA2
				\cr
				\hline
				\hline
				$\mathcal{R}$ & - & - & 62.3 & 34.9 & 66.4 \cr
				\hline
				\multirow{9}{*}{$\mathcal{E}$} & 8 & 2 & 60.6 & 27.0  & 46.9 \cr
				
				& 8 & 3 & 64.2 & 27.8  & 61.8 \cr
				
				& 8 & 4 & 58.5 & 27.2  & 62.3 \cr
				
				& 12 & 2 & 62.5 & 33.1  & 68.2 \cr
				
				& 12 & 3 & 68.7 &  35.8 & 65.1 \cr
				
				& 12 & 4 & 65.5 &  34.5 & 67.4 \cr
				
				& 16 & 2 & \textbf{70.4} & \textbf{36.9} & \textbf{70.6} \cr
				
				& 16 & 3 & 68.3 & 36.7 & 68.3 \cr
				
				& 16 & 4 & 69.1 & 34.8 & 69.3 \cr
				\bottomrule[1.5pt]
		\end{tabular}}
	\end{center}
	\label{episode}
\end{table}

\textbf{Training method analysis.} An episode-based training method is used in our experiments to make the model gain better generalization ability. For each mini-batch, we sample $M$ categories and $N$ images for each category. We vary the value of $M$ with $\{8, 12, 16\}$ and the value of $N$ with $\{2, 3, 4\}$, and observe the \textbf{H} under these values. To further analyze the performance of the episode-based training method, we compare its performance with the random sampling training method with a mini-batch of 64. Table~\ref{episode} shows that the episode-based training method has better performance than the random sampling training method. The model can be generalized to the recognition of all categories (seen and unseen categories) only by learning the seen categories. When $M=16$ and $N=2$, the model can get the highest accuracy.

\begin{figure*}[t]
	\centering
	\includegraphics[width=1\linewidth]{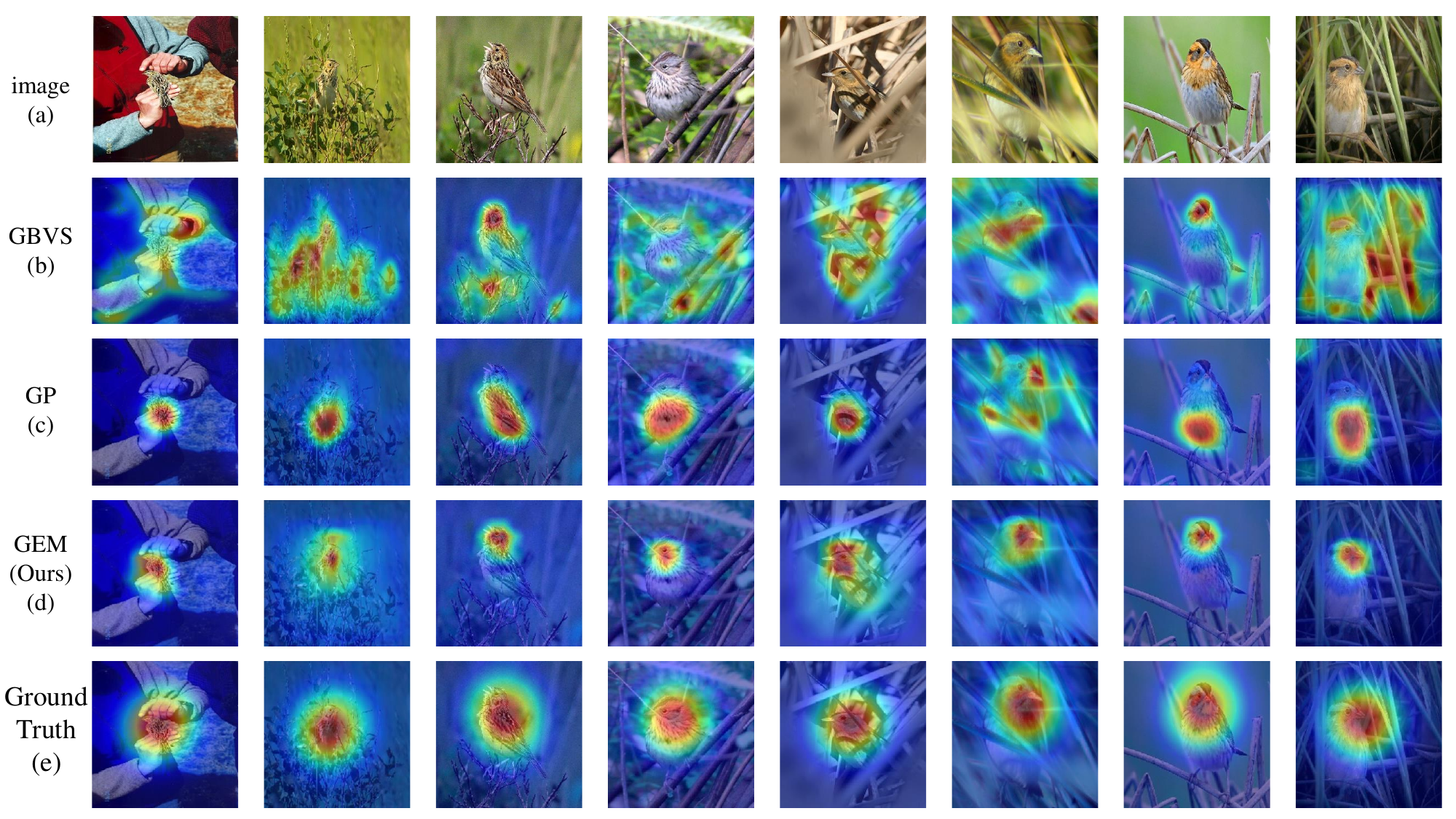}
	\caption{Visualization of predicted gaze maps of randomly selected unseen images on CUB-VWSW from different gaze estimation methods.}
	\label{gazemap}
\end{figure*}

\subsection{GEM Evaluation}

\textbf{Quantitative results.} We use two standard evaluation metrics for gaze estimation: Area Under the Curve (AUC)~\cite{borji2013analysis} and Normalized Scanpath Saliency (NSS)~\cite{peters2005components}. AUC is the area under a curve of true positive rate versus false positive rate for different thresholds on the estimated gaze map. It is a commonly used evaluation metric in saliency prediction. NSS is a simple correspondence measure between saliency maps and ground truth, computed as the average normalized saliency at fixated locations. We compared our GEM with two gaze estimation baselines include: bottom-up saliency prediction method Graph Based Visual Saliency (GBVS)~\cite{harel2007graph} and top-down task-specific Gaze Prediction (GP)~\cite{huang2018predicting}.
Table~\ref{gazeresult} shows the results of different methods for gaze estimation on the unseen images of CUB-VWSW dataset. We can see that our GEM achieves better gaze estimation results.

\begin{table}[t]
	\centering
	\caption{Results of different methods for gaze estimation on unseen images of CUB-VWSW dataset.}
	\begin{center}
	\setlength{\tabcolsep}{7mm}
		\begin{tabular}{c|c|c}
			\toprule[1.5pt]
			Methods&AUC&NSS\cr
			\hline
			 GBVS~\cite{harel2007graph}& 0.793  & 1.003 \cr
			 GP~\cite{huang2018predicting}& 0.836  & 1.430  \cr
			 GEM (Ours)& \textbf{0.914} & \textbf{2.244} \cr
			\bottomrule[1.5pt]
		\end{tabular}
	\end{center}
	\label{gazeresult}
\end{table}

\textbf{Qualitative analysis.} Figure~\ref{gazemap} shows the qualitative results of different gaze estimation methods on randomly selected unseen images from CUB-VWSW dataset. Our GEM can detect more accurate attribute regions that humans pay more attention to compared with both the bottom-up method GBVS and top-down method GP. These regions are helpful to distinguish different categories. Because our model is zero-shot learning targeted, GEM can learn more refined category attribute areas. In the visualization results, the predicted gaze maps are more concentrated than ground truth maps. In general,
these results demonstrate that the GEM plays an important role in improving the accuracy of attribute parts prediction.

%------------------------------------------------------------------------
\section{Conclusion}

In this paper, a novel goal-oriented gaze estimation method has been introduced to improve ZSL tasks. To mimic human cognitive process with unseen classes, the GEM learns the discriminative attributes with semantic query-guided attention and real human gaze (if available) supervision. Finally, the localized discriminative attributes improve the global image feature representation for ZSL. Our GEM-ZSL achieves superior or competitive performance on three ZSL benchmarks which demonstrates the effectiveness of the discriminative attributes learned by the GEM. Our work shows the promising benefits of collecting human gaze dataset and automatic gaze estimation algorithms on computer vision tasks. For the future work, further investigation on the discriminative attribute localization for ZSL is an intrinsic and important direction.

%\noindent\textbf{Acknowledgement} This work was supported by National Natural Science Foundation of China

{\small
	\bibliographystyle{ieee_fullname}
	\bibliography{egbib}
}

\end{document}